\documentclass[lettersize,journal]{IEEEtran}
\usepackage[bookmarks=false]{hyperref}
    \hypersetup{colorlinks,
      linkcolor=blue,
      citecolor=blue,
      urlcolor=blue}

\usepackage{float}
\usepackage{algorithm}
\usepackage{algorithmicx}
\usepackage{algpseudocode}
\usepackage{times}
\usepackage{subcaption}
\usepackage{setspace}
\usepackage{enumitem}
\usepackage{times}
\usepackage{epsfig}
\usepackage{graphicx}
\usepackage{amsmath}
\usepackage{amssymb}
\usepackage{textcomp}
\usepackage{multirow}
\usepackage{adjustbox}
\usepackage[utf8]{inputenc}
\usepackage{gensymb}
\usepackage{flushend}
\usepackage{xcolor}
\usepackage{colortbl}
\usepackage{hhline}
\usepackage{verbatim}
\usepackage{amssymb}

\usepackage[figuresright]{rotating}

\title{Modified RRT* for Path Planning in Autonomous Driving}

\author{Sugirtha T$^{a}$, 
Pranav S$^{b}$, 
Nitin Benjamin Dasiah$^{b}$, 
Sridevi M$^{b}$ \\

{$^{a}$Department of CSE, Indian Institute of Information Technology, Kottayam, India \\
$^{b}$Department of CSE, National Institute of Technology, Tiruchirappalli-620015, India}
}

\begin{document}
\maketitle

\begin{abstract}
Essential tasks in autonomous driving includes environment perception, detection and tracking, path planning and action control. This paper focus on path planning, which is one of the challenging task as it needs to find optimal path in highly complex and dynamic environments. Usually, a driving scenario has large num- ber of obstacles in their route. In this paper, we propose a two-stage path planning algorithm named Angle-based Directed Rapidly exploring Random Trees (AD- RRT*) to address the problem of optimal path in complex environment. The proposed algorithm uses A* algorithm for global path planning and modifies RRT* to bound the samples using angle. The efficiency of the proposed algorithm is evaluated through experiments in different scenarios based on the location and number of obstacles. The proposed algorithm showed higher rate of convergence with reduced time and less number of nodes than the base RRT* algorithm.
\end{abstract}









\section{Introduction}\label{sec1}
Autonomous driving has emerged as a hot research topic in recent decades. Hence, many researchers and automotive companies spend billions of dollars in developing various modules to make Level-5 self driving cars. Deep learning has achieved state-of-the-art results in both semantic and geometric tasks in autonomous driving which includes object detection~\cite{klingner2023x3kd, mohapatra2021bevdetnet, rashed2020fisheyeyolo, 10.1117/1.JEI.32.1.011004}, semantic segmentation \cite{briot2018analysis, chennupati2019auxnet, sistu2019real, siam2017convolutional}, depth prediction \cite{kumar2018near, kumar2021svdistnet, sekkat2022synwoodscape, kumar2020unrectdepthnet}, adverse weather detection \cite{dhananjaya2021weather, uricar2019desoiling, uricar2019challenges, ramachandran2021woodscape}, moving object detection \cite{siam2018modnet,  rashed2019motion, mohamed2021monocular, kiran2011improved}, SLAM \cite{tripathi2020trained, yahiaoui2019overview, yahiaoui2011impact},  multi-task learning \cite{leang2020dynamic,  sistu2019neurall, rashed2019optical, kumar2023surround} and sensor fusion \cite{popperli2019capsule,  dasgupta2022spatio, eising2021near}. The preliminary task is object detection and tracking, and the next crucial task in a self-driving car is to find the optimal path from source to destination point. Path planning helps the ego-vehicle to navigate through the environment by considering the surrounding obstacles.\\
\indent Path planning algorithms have widespread applications in autonomous driving \cite{Lan2015ContinuousCP}, planetary and space missions \cite{Lau2014RealTimePP}, Unmanned Aerial Vehicles (UAVs) \cite{article3} , computerized robotic surgery \cite{10.1007/978-3-642-33415-3_58}, , artificial intelligence for video games \cite{naderi2015rt}, and molecular biology \cite{reif1999nano}.\\
\indent Path planning refers to generation of collision free trajectory from an initial state to destination state and a velocity profile to the controller by \ considering \ the \ following factors :
\begin{itemize}
    \item State of the self driving car which includes velocity, inclination etc.,
    \item Surrounding environment of the ego-vehicle which includes static and dynamic obstacles, drivable spaces
    \item Traffic laws and allowed maneuvers
\end{itemize}

The \ path \ planning \ task \ is \ accomplished \ as \ a \ sequence \ of four subtasks \ namely (i) \ Global \ path \ planning \ (ii) behavioural planner (iii) local path planning (iv) local feedback control.\\
\indent Global path planning refers to finding the collision free optimal path from starting state to the destination state. Local path planning refers to planning the intermediate local move by creating the nodes within the coordinate space.

Various algorithms were proposed over past four decades to solve path-planning problem in static and dynamic environment which includes grid based \cite{4082128,article1}, Artificial Potential Field (APC) \cite{6722915,Hwang1992GrossMP,Goerzen2010ASO}, evolutionary methods \cite{10.1016/j.advengsoft.2014.09.006, doi:10.5772/63484, KALA20123817, ZHU2014153}, geometric and neural networks \cite{10.1016/j.neucom.2013.07.055, Zhu2014ThePP}. An extensive survey that describes the merits and demerits of these path planning algorithms can be found in \cite{Nosrati2012InvestigationOT, inproceedings1,Noreen2016OptimalPP}.
\begin{figure}[!hbt]
\begin{center}
\includegraphics[height=4cm,width=6cm]{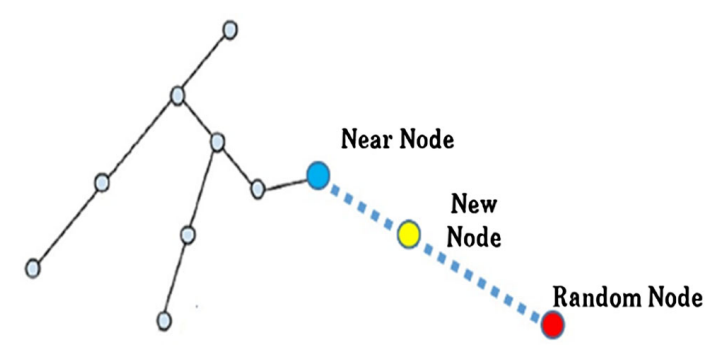}
\caption{Tree expansion process of RRT* \cite{Karaman2011SamplingbasedAF}}
\label{fig-tree}
\end{center}
\end{figure}
Though, there were several techniques employed for the path planning, the following issues were identified: 
\begin{itemize}
    \item Slow convergence to optimal solution
    \item Longer paths from source to destination
    \item High memory requirement by creating large number of nodes
    \item Not suitable for complex dynamic environment
\end{itemize}
To address these issues, this paper proposes a variant of RRT*. The major contributions of this paper is given as follows :
\begin{itemize}
\item Proposed an angle based detector in front of the ego-vehicle to quickly find optimal path
\item Produce comparatively shorter path than base RRT*
\item Consumes less memory
\item Deals with complex environment
\end{itemize}
The remaining section of the paper is organized as follows : Section 2 describes the detailed explanation of various existing path planning algorithms. Section 3 explains the proposed Angle-based Directed RRT* in detail manner. The experimental results and performance analysis of the proposed method is provided in Section 4. Finally, Section 5 concludes the paper. 

\section{Related Work}
This section explains the local path planning methods exists in the literature. We classify them into two categories namely, (i) Potential field based path planning methods and (ii) Graph based path planning methods.\\
\indent Conventional path planning approaches use Potential Field (PF) \cite{Castañeda08} which includes attractive and repulsive potential fields. The vehicle is attracted towards the goal by the attractive field and repelled from obstacles by the repulsive field. The limitation of conventional Potential Field Method (PFM) is that it gets stuck at local minima. To overcome the limitation of conventional PFM, modified Artificial Potential Field (APF) \cite{7995717} was proposed where the repulsive field makes use of goal co-ordinates from camera, obstacle co-ordinates from radar and autonomous vehicle co-ordinate from Global positioning system (GPS) and computes the improved artificial potential field. Obstacle dependent Gaussian Potential Field (GPF) proposed in \cite{Cho2018ARO} calculates repulsive potential field by considering the objects information from the sensor within a threshold and uses yaw angle information and calculate attractive field to find the path.\\
\indent In recent years, Sampling Based Planning (SBP) methods are probabilistic complete and they are most extensively used for finding the optimal path. They perform random sampling in search space and are computationally efficient in solving high dimensional real time complex problems. They create a roadmap of possible trajectories in configuration space without explicitly using the obstacle information.\\ 
\begin{figure*}[hbt!]
\begin{center}
\includegraphics[height=7cm, width=6cm]{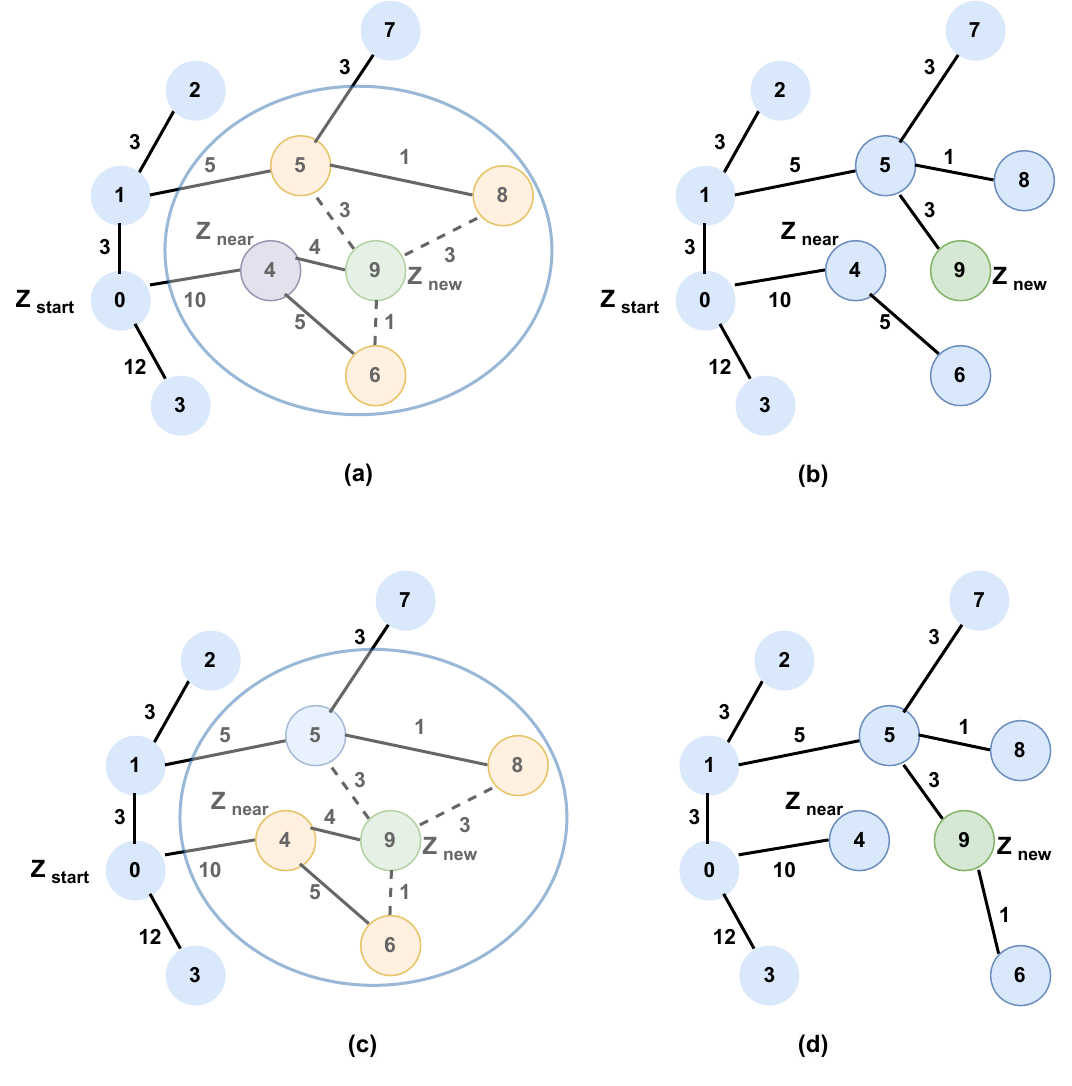}
\caption{Rewiring Process of RRT* \cite{Karaman2011SamplingbasedAF}. (a) Finding near vertices (b) Selection of best parent (c) cost check (d) rewiring with minimum cost }
\label{fig-rewire}
\end{center}
\end{figure*}
\begin{figure*}[!hbt]
    \centering
    \includegraphics[height=5cm, width=12cm]{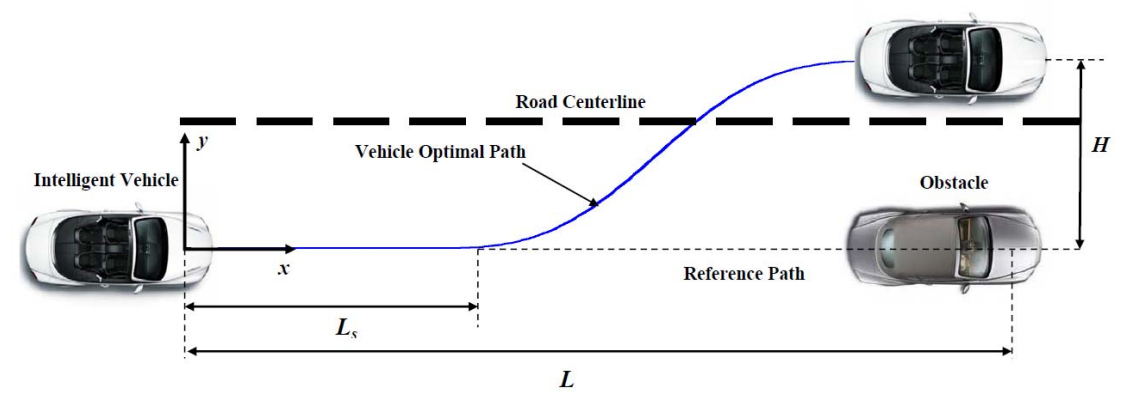}
    \caption{Lane changing maneuver \cite{mashadi2014global}}
    \label{fig:W-3}
\end{figure*}
\indent A study and analysis of various SBP techniques can be found in \cite{6722915}. Probabilistic Road Map (PRM) \cite{508439} based approach is widely used in highly structured static environment and suitable for holonomic constraints. But, they depend on geometry of obstacles and doesn’t ensure asymptotic optimality.  Rapidly exploring Random Tree (RRT) is one of the fastest path planning algorithms which supports dynamic cluttered environment and can be extended to non-holonomic constraints. But, it does not ensure asymptotic optimality.
Due to its advantages, various extensions of RRT were proposed to improve the efficiency. A major breakthrough was witnessed in high dimensional optimal path planning algorithms with asymptotic optimality after the introduction of RRT* \cite{Karaman2011SamplingbasedAF} in 2010. The major advantage of RRT* is the quick exploration of initial path and optimizes it in successive iterations. The asymptotic optimal property of RRT* generates near optimal path when the number of iterations approach infinity makes it very expedient for real time applications. The tree expansion process of RRT* is shown in Fig \ref{fig-tree}. Nearest neighbour search and rewiring operations are shown in Fig \ref{fig-rewire}. However, RRT* never produce the optimal path in finite time and the convergence rate is slow. This happens because it explores and samples the nodes in whole configuration space. Also, the space complexity is high as the number of nodes in the tree is high.\\
\indent To address the above drawbacks, a new variant of RRT*, named RRT*-smart \cite{doi:10.5772/56718} was proposed. It uses intelligent sampling and path optimization for faster convergence of optimal path. RRT*-smart finds the initial path similar to RRT* and optimizes the path by performing intelligent sampling at regular intervals using biasing ratio. RRT*-smart converges faster than RRT*, but adjusting the biasing ratio causes computational overhead. RRT*-Adjustable Bounds (RRT*-AB) address the issues of RRT*-smart such as dense tree and high memory requirements. RRT*-AB finds optimal path by exploring only favorable regions and focusing on features like (i) connectivity (ii) intelligent bounded sampling and (iii) path optimization. In order to overcome the limitations of RRT and RRT*, we propose a two-stage path planning algorithm for autonomous driving by modifying base RRT*.\\
\indent In this paper, we focus on the path planning task. To ensure safety, the surrounding vehicles need to follow correct lane, overtake other vehicles when it is safe to do so, stop at the intersections and take U-turns if required and do efficient parking in the allotted space. A lane changing maneuver scenario of an autonomous vehicle is shown in Fig \ref{fig:W-3}.\\

\section{Proposed Work}
\begin{figure*}[hbt!]
\begin{center}
\includegraphics[height=4cm, width=7cm]{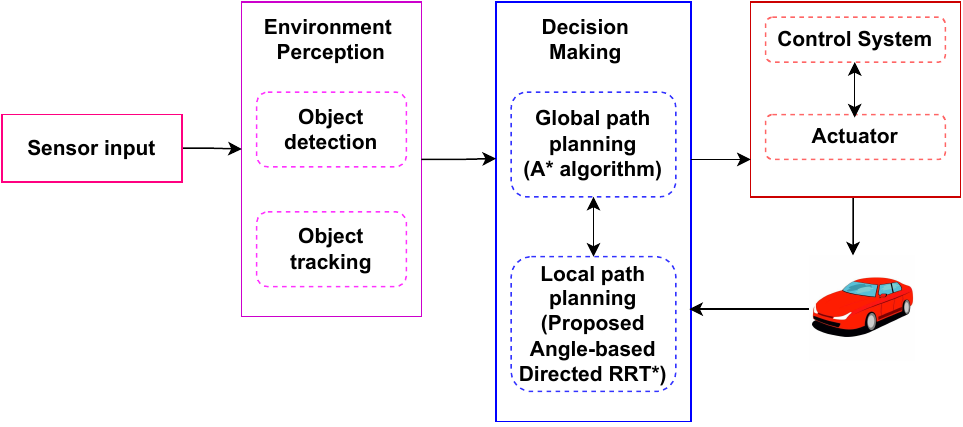}
\caption{General pipeline of autonomous driving tasks [focus on Decision making task]}
\label{fig-pp}
\end{center}
\end{figure*}

This section explains the proposed two-stage path planning algorithm which finds global path and local path to plan the maneuvers. The overall pipeline of various tasks involved in autonomous driving is shown in Fig \ref{fig-pp}. The proposed work focus on the decision making task which finds the optimal path from source to destination points using \textbf{Algorithm 1}.\\

The proposed algorithm utilizes A* algorithm \cite{8813752} to create a global path for deciding which roads to turn. This global path will be used by the local path to decide the angles. RRT and RRT* algorithms are generally very inefficient for real-time path planning. They tend to search large portions of the map, that is not necessarily useful in case of vehicles. We propose Angle-Based Directed RRT* (AD-RRT*) to not only make the algorithm more efficient for vehicles, but to add another layer of algorithm on top of it, so the algorithm can be tuned to required situations with the help of hyper-parameters. The proposed method uses angle to bound the sample. As the time taken to find a path increases, the angle increases with expansion of the search area. The angle uses the global path for the initial angle. If there is an obstacle and the algorithm is unable to find a path, it expands its search angle, equally on both sides.
After a local path reaches a given distance towards the global path, the vehicle will move towards it, and updation of the point from which the angle is calculated, reducing the search area back to 0\degree.  For most cases, it will force the vehicle to go towards the goal path reaching the final state. This method of bounding the search area, will not always produce an optimal path and in worst case scenarios take more time to search for path. Therefore, the proposed algorithm is best used for vehicles and needs more optimizations to work efficiently. It is required to use visibility regions to search in shorter distances than longer distances when there are more crowd. This allows for searching through narrow areas if needed.\\

\noindent \textbf{Notation :}\\
\noindent Let $\mathbb{Z} \subset \mathbb{R}^{n}, n \in \mathbb{N}$ represents the configuration space and $\mathbb{Z}_{obs}$ represents the obstacle region and $\mathbb{Z}_{\mathrm{free}}=\mathbb{Z} / \mathbb{Z}_{\mathrm{obs}}$ denotes the obstacle free region.  $\mathbb{Z}_{src}$ and $\mathbb{Z}_{destn}$ denotes the starting and destination points respectively. $u:[0: s]$ denotes the path connecting the states $z_{1}$ and $z_{2}$ such that $u(0)=z_{1}$ and $u(s)=z_{2}$, where s is the positive scalar length of the path. $\Sigma_{f}$ denotes the set of paths in $\mathbb{Z}_{\mathrm{free}}$ that are collision free and $\sigma_{f} \in Z_{\text {free }}$ denotes end to end feasible path. ${\mathbb{C}_{\text {Region }}}$ denotes 
a region with connected nodes near to the starting node $\mathbb{Z}_{src}$.\\

\begin{algorithm*}[!hbt]
\caption{Angle based Directed RRT*}\label{alg:cap}
\begin{algorithmic}[1]
\State $T \gets \text { InitializeTree }() $
\State $T \gets \text { InsertNode }\left(\phi, z_{\text {src }}, T\right)$ 
\State $Angle \gets 0\degree$
\State $A_{Length} \gets collect from previous layer visibility$
\State $C_{\text {Region }} \gets \text { ConnectivityRegion }\left(z_{\text {src }}, A_{Length}, A_{angle}, T\right)$ 
\For {$i \gets  0 \quad \text {to} \quad N $} 

\State $z_{\text {rand }} \gets \text { Bounded Sample }\left(i, C_{\text {Region }}\right)$

\State $z_{\text {nearest }} \gets \text { Nearest }\left(T, z_{\text {rand }}\right)$

\State $\left(z_{\text {new }}, U_{\text {new }}\right) \gets \text { Steer }\left(z_{\text {nearest }}, z_{\text {rand }}\right)$
\If
{$ \text{CollisionCheck }\left(z_{\text {new }}\right)$}
\State $z_{\text {near }} \gets \operatorname{Near}\left(T, z_{\text {new }},|V|\right)$
\State $z_{\text {min }} \gets \text { ChooseParent }\left(z_{\text {near }}, z_{\text {nearest }}, z_{\text {new }}\right)$
\State $T \gets \text { Insert Node }\left(z_{\min }, z_{\text {new }}, T\right)$
\State $T \gets \text { Rewire }\left(T, z_{\text {near }}, z_{\min }, z_{\text {new }}\right)$
\If
{$PathFound(T)$}
\State $C_{\text {Region }} \leftarrow \text { ConnectivityRegion }\left(z_{\text {init }}, z_{\text {destn }}, T\right)$
\Else 
\quad \textbf{if} \quad {$ Timetakenhigh \left(\right)$}
\State $C_{\text {Region }} \gets \text { ConnectivityRegion }\left(z_{\text {init }}, z_{\text {goal }}, T\right)$
\State $angle \gets angle + increase$
\State $C_{\text {Region }} \gets \text { ConnectivityRegion }\left(z_{\text {src }}, A_{Length}, A_{angle}, T\right)$
\State $T \gets \text { Prune Path }(T)$ 
\State \Return
\EndIf
\EndIf
\EndFor
\end{algorithmic}
\end{algorithm*}
\begin{table*}[!hbt]
\caption{Comparison of base RRT* with Proposed Angle-based Directed RRT*}
\label{tab:pp}
\resizebox{.70\textwidth}{!}{\begin{minipage}{\textwidth}
\begin{tabular}{|c|cccc|cccc|}
\hline
\multirow{2}{*}{\textbf{No. of obstacles}} & \multicolumn{4}{c|}{\textbf{Base RRT*}}                                                                                                                                                                           & \multicolumn{4}{c|}{\textbf{Angle based Directed   RRT* (AD-RRT*)}}                                                                                                                                              \\ \cline{2-9} 
                                           & \multicolumn{1}{c|}{\textbf{No. of nodes}} & \multicolumn{1}{c|}{\textbf{\begin{tabular}[c]{@{}c@{}}Time taken \\ (s)\end{tabular}}} & \multicolumn{1}{c|}{\textbf{Total path cost}} & \textbf{Average path cost} & \multicolumn{1}{c|}{\textbf{No. of nodes}} & \multicolumn{1}{c|}{\textbf{\begin{tabular}[c]{@{}c@{}}Time taken\\ (s)\end{tabular}}} & \multicolumn{1}{c|}{\textbf{Total path cost}} & \textbf{Average path cost} \\ \hline
50                                         & \multicolumn{1}{c|}{8174}                  & \multicolumn{1}{c|}{21.28}                                                              & \multicolumn{1}{c|}{41203.74}                 & 934.01                     & \multicolumn{1}{c|}{6904}                  & \multicolumn{1}{c|}{20.7}                                                              & \multicolumn{1}{c|}{39556.83}                 & 791.13                     \\ \hline
68                                         & \multicolumn{1}{c|}{9070}                  & \multicolumn{1}{c|}{30.05}                                                              & \multicolumn{1}{c|}{39226.16}                 & 784.52                     & \multicolumn{1}{c|}{8534}                  & \multicolumn{1}{c|}{31.42}                                                             & \multicolumn{1}{c|}{38502.22}                 & 770.04                     \\ \hline
70                                         & \multicolumn{1}{c|}{11803}                 & \multicolumn{1}{c|}{78.74}                                                              & \multicolumn{1}{c|}{46700.4}                  & 934.01                     & \multicolumn{1}{c|}{8656}                  & \multicolumn{1}{c|}{67.18}                                                             & \multicolumn{1}{c|}{43174.01}                 & 863.48                     \\ \hline
80                                         & \multicolumn{1}{c|}{15353}                 & \multicolumn{1}{c|}{59.5}                                                               & \multicolumn{1}{c|}{44426.36}                 & 888.52                     & \multicolumn{1}{c|}{10461}                 & \multicolumn{1}{c|}{47.55}                                                             & \multicolumn{1}{c|}{41802.47}                 & 836.04                     \\ \hline
85                                         & \multicolumn{1}{c|}{13198}                 & \multicolumn{1}{c|}{56.48}                                                              & \multicolumn{1}{c|}{44419.73}                 & 888.39                     & \multicolumn{1}{c|}{11834}                 & \multicolumn{1}{c|}{56.27}                                                             & \multicolumn{1}{c|}{41530.99}                 & 830.62                     \\ \hline
100                                        & \multicolumn{1}{c|}{16785}                 & \multicolumn{1}{c|}{99.72}                                                              & \multicolumn{1}{c|}{47707.62}                 & 954.15                     & \multicolumn{1}{c|}{12567}                 & \multicolumn{1}{c|}{88.23}                                                             & \multicolumn{1}{c|}{44661.59}                 & 893.23                     \\ \hline
\end{tabular}
\end{minipage}}
\end{table*}
The proposed algorithm finds a feasible path $\sigma_{f} \in \Sigma_{f}$ from $\sigma_{f(0)}=z_{\text {src }}$ to $\sigma_{f(s)}=z_{\text {destn}}$ by building a Tree $T=(V, E)$, that has the vertices ($V$) sampled from obstacle free region $\mathbb{Z}_{\mathrm{free}}$ and ($E$), the edges that connect these vertices.\\ 
\begin{figure*}[!hbt]
\begin{center}
\includegraphics[width=5cm,height=7cm]{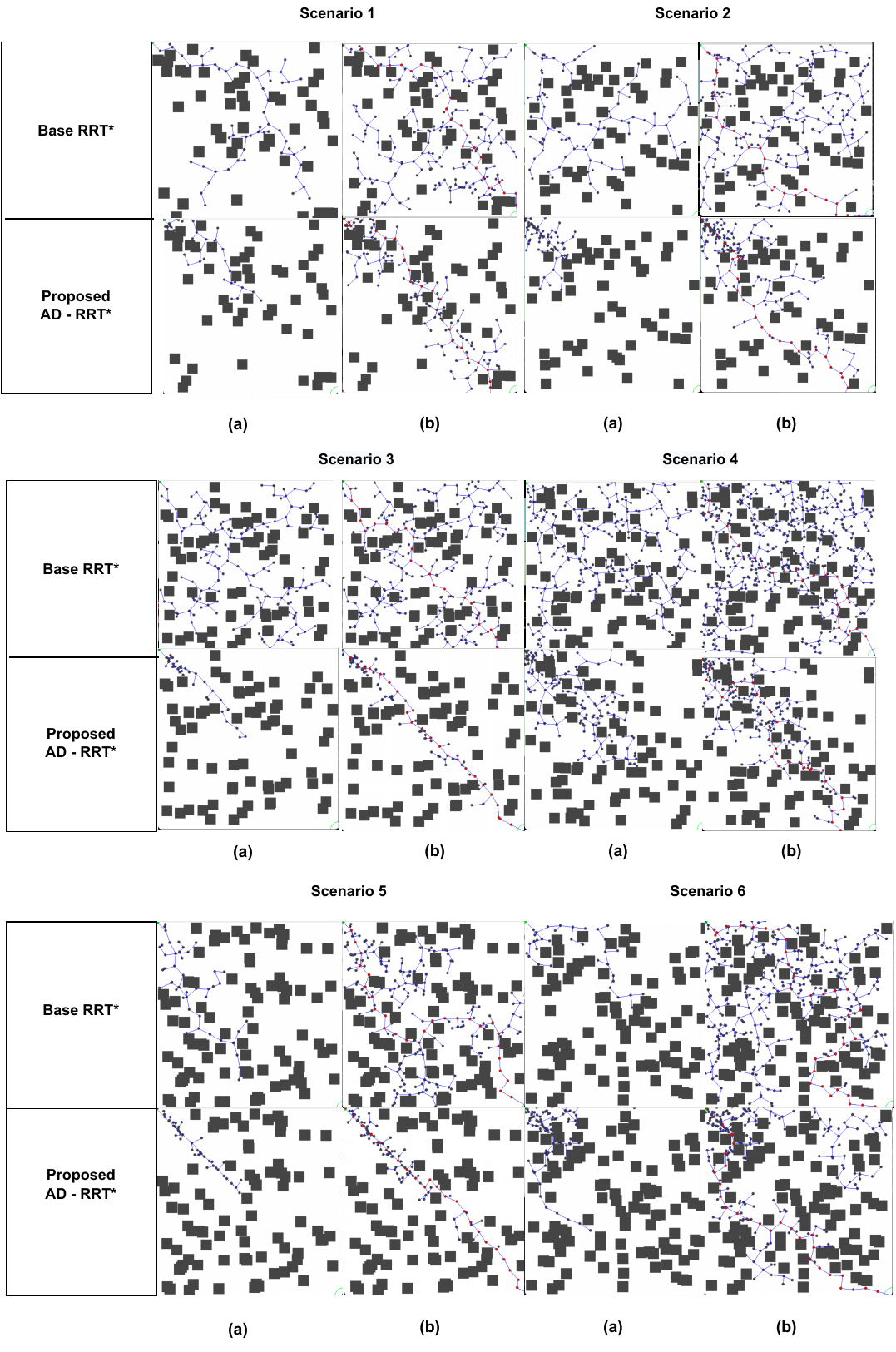}
\caption{Visual results of Path planning for Scenarios S1, S2, S3, S4, S5 and S6. (a) and (b) under each scenario denotes the intermediate and final step in path planning.}
\label{fig-output-1}
\end{center}
\end{figure*}
\begin{figure*}[!hbt]
\begin{center}
\includegraphics[width=7cm,height=7cm]{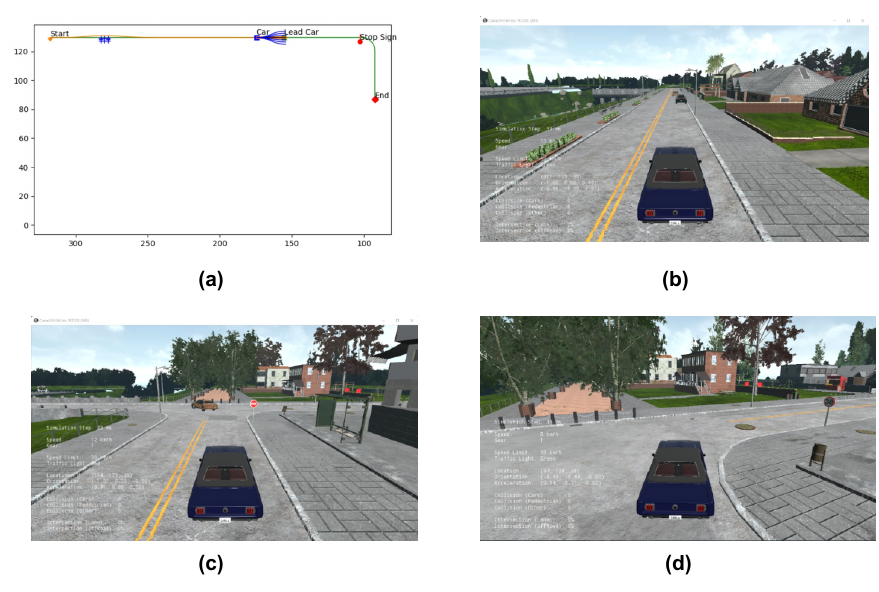}
\caption{Visual results of proposed method on CARLA Simulator (a) Plot of global path planning, Local path planning - (b) Initial step (c) Intermediate step and (d) Final step where the ego-vehicle takes the right turn}
\label{fig-carla}
\end{center}
\end{figure*}
The major procedures of the algorithm uses the following task :\\

\textbf{Bounded-sampling :} A state ${Z_{rand}}$is placed randomly from obstacle free region $\mathbb{Z}_{free}$.\\

\textbf{Connectivity Region :} A connectivity region ${\mathbb{C}_{\text {Region }}}$ is identified in search space near to $\mathbb{z}_{src}$ using the area with length.\\

\indent \textbf{Nearest :} The nearest node from $T=(V, E)$ to ${z_{rand}}$ is identified by $Nearest(T, z_{rand})$ based on the cost computed by the distance function as given in equation (\ref{eq:dis}).
\begin{equation}
\Delta d=\sqrt{\left(x_{2}-x_{1}\right)^{2}+\left(y_{2}-y_{1}\right)^{2}} 
\label{eq:dis}
\end{equation}

\textbf{Steer :} The procedure $\operatorname{Steer}\left(z_{\text {near }}, z_{\text {rand }}\right)$ provides a control input $u[0, T]$ that drives the system from $x(0)=z_{\text {rand}}$ to $x(T)=z_{\text {near}}$ along the path $x:[0, T] \rightarrow X$ giving $z_{\text {new}}$ at a distance $\Delta q$ from $z_{\text {near}}$ towards $z_{\text {rand}}$, where $\Delta q$ is the incremental distance.

\textbf{Collision check :} This procedure determines if a path $u[0, T]$ exists in the obstacle free region $\mathbb{Z}_{free}$ for all $t=0$ to $t=T$.\\

\textbf{Near :} This procedure returns the nearby neighbouring nodes as given by equation (\ref{eq:1}):
\begin{equation}
\label{eq:1}
k=\gamma(\log (n) / n)^{(1 / d)}
\end{equation} 

\textbf{Choose parent :} It selects the best parent $\mathbb{z}_{min}$ which gives the minimum cost from the source point to $\mathbb{z}_{new}$.\\

\textbf{Insert node :} It adds a node $\mathbb{z}_{min}$ to $V$ in the tree $T=(V,E)$ and connects it to $\mathbb{z}_{min}$ as its parent.\\

\textbf{Rewire :} Rewiring is performed after the parent is selected. It computes the cost of all nodes through $\mathbb{z}_{new}$ and compared with their previous costs. If the cost is minimum for any node, then $\mathbb{z}_{new}$ will be made as its new parent.\\

\textbf{Prunepath :} It returns the nodes that are connectable from source to destination point in the obstacle free region.  A tree $T$ is initialized with $\mathbb{z}_{src}$ as the root node. Bounded sampling randomly selects nodes restricted by an angle from ${\mathbb{C}_{\text {Region }}}$ and the tree is populated with these nodes. ${\mathbb{C}_{\text {Region }}}$ is defined by expansion distance scale in a search space between $\mathbb{z}_{src}$ and $\mathbb{z}_{destn}$ using equation (\ref{eq:2}):
  \begin{equation}
  \label{eq:2}
D_{\text {scale }}=E / m
\end{equation}
where,
\indent \quad E – size of the environment map\\
\indent \quad \quad \quad \quad m – expansion factor\\
\noindent
When a single scan of ${\mathbb{C}_{\text {Region }}}$ is completed using bounded sampling and no path is found in the first scan and the time exceeds the limit, then the angle of ${\mathbb{C}_{\text {Region }}}$ is increased as mentioned in Steps 15-19 of the Algorithm 1.\\

\section{Experimental results and Analysis}
This section demonstrates the simulation results of proposed Angle-Based Directed RRT* with CARLA simulator \cite{Dosovitskiy17}. The experiments were carried out with different scenario based on the density (cluster) of the obstacle nodes. The proposed approach is implemented using Python in an Intel i7, NVIDIA 1080 GTX GPU. \\ 

The proposed algorithm is experimented on six different scenarios i.e. ( S1, S2, S3, S4, S5 and S6) which represents the distribution of obstacle nodes (N). The scenarios are explained as follows :\\
\begin{enumerate}
    \item S1 – Obstacles densed near starting point with N=50
    \item S2 - Sparse distribution of obstacles throughout
the path from starting to destination point with N=50
    \item S3 - Sparse distribution of obstacles throughout
the path from starting to destination point with N=68
    \item S4 – Entire region equally densed with obstacles N=70
    \item S5 - Obstacles densed in middle region with N=80
    \item S6 - Obstacles densed near starting, middle and destination region with N=85
\end{enumerate}

Fig \ref{fig-output-1} shows the path planning results of base RRT* and the proposed Angle-Based Directed RRT* under different scenarios. The global path planning A* algorithm finds the path between $Z_{src}$ and $Z_{destn}$. The local path planning is performed by the proposed Angle-Based Directed RRT*. From the Figure, it is clear that the base RRT* expands the tree in entire search space and thus results in generating dense tree with more number of unnecessary nodes. On the contrary, the proposed AD-RRT* restricts the search space to an angle and finds the optimal path by generating less number of nodes i.e the convergence rate is high with less time.\\ 

Table \ref{tab:pp} shows the comparison between base RRT* and proposed AD-RRT* for varying number of obstacles. The proposed method finds optimal path in less time by generating less number of nodes. Also, the total path cost and average path cost are less compared to base RRT*. Fig \ref{fig-carla} shows the output of CARLA simulator.\\

\section{Conclusion}
In this paper, a two-stage path planning algorithm called Angle-based Directed RRT* is proposed to find the optimal path for autonomous vehicles in complex driving environments. The proposed method deployed A* algorithm to find global path and compute the local path in restricted search space by modifying RRT*. Analysis of the proposed algorithm is carried out in six different scenarios using CARLA simulator and the results show faster convergence with less number of nodes than the base RRT* algorithm. In future, the proposed algorithm is extended to dynamic environment and the performance is analysed.

\section{Acknowledgement}
This publication is an outcome of the R\&D work undertaken in the project under TiHAN Faculty Fellowship of NMICPS Technology Innovation Hub on Autonomous Navigation Foundation being implemented by Department of Science \& Technology National Mission on Interdisciplinary Cyber-Physical Systems (DST NMICPS) at IIT Hyderabad.

\bibliographystyle{elsarticle-num}
\bibliography{ref} 

\begin{thebibliography}{10}
\expandafter\ifx\csname url\endcsname\relax
  \def\url#1{\texttt{#1}}\fi
\expandafter\ifx\csname urlprefix\endcsname\relax\def\urlprefix{URL }\fi
\expandafter\ifx\csname href\endcsname\relax
  \def\href#1#2{#2} \def\path#1{#1}\fi

\bibitem{klingner2023x3kd}
M.~Klingner, S.~Borse, V.~R. Kumar, B.~Rezaei, V.~Narayanan, S.~Yogamani, F.~Porikli, X3kd: Knowledge distillation across modalities, tasks and stages for multi-camera 3d object detection, in: Proceedings of the IEEE/CVF Conference on Computer Vision and Pattern Recognition, 2023, pp. 13343--13353.

\bibitem{mohapatra2021bevdetnet}
S.~Mohapatra, S.~Yogamani, H.~Gotzig, S.~Milz, P.~Mader, Bevdetnet: bird's eye view lidar point cloud based real-time 3d object detection for autonomous driving, in: 2021 IEEE International Intelligent Transportation Systems Conference (ITSC), IEEE, 2021, pp. 2809--2815.

\bibitem{rashed2020fisheyeyolo}
H.~Rashed, E.~Mohamed, G.~Sistu, V.~R. Kumar, C.~Eising, A.~El-Sallab, S.~Yogamani, Fisheyeyolo: Object detection on fisheye cameras for autonomous driving, in: Proceedings of the Machine Learning for Autonomous Driving NeurIPS 2020 Virtual Workshop, Virtual, Vol.~11, 2020.

\bibitem{10.1117/1.JEI.32.1.011004}
S.~Thayalan, S.~Muthukumarasamy, K.~Santhakumar, K.~B. Ravi, H.~Liu, T.~Gauthier, S.~Yogamani, {Monocular three-dimensional object detection using data augmentation and self-supervised learning in autonomous driving}, Journal of Electronic Imaging 32~(1) (2022) 011004.
\newblock \href {http://dx.doi.org/10.1117/1.JEI.32.1.011004} {\path{doi:10.1117/1.JEI.32.1.011004}}.

\bibitem{briot2018analysis}
A.~Briot, P.~Viswanath, S.~Yogamani, Analysis of efficient cnn design techniques for semantic segmentation, in: Proceedings of the IEEE Conference on Computer Vision and Pattern Recognition Workshops, 2018, pp. 663--672.

\bibitem{chennupati2019auxnet}
S.~Chennupati, G.~Sistu, S.~Yogamani, S.~Rawashdeh, Auxnet: Auxiliary tasks enhanced semantic segmentation for automated driving, arXiv preprint arXiv:1901.05808.

\bibitem{sistu2019real}
G.~Sistu, I.~Leang, S.~Yogamani, Real-time joint object detection and semantic segmentation network for automated driving, arXiv preprint arXiv:1901.03912.

\bibitem{siam2017convolutional}
M.~Siam, S.~Valipour, M.~Jagersand, N.~Ray, S.~Yogamani, Convolutional gated recurrent networks for video semantic segmentation in automated driving, in: 2017 IEEE 20th International Conference on Intelligent Transportation Systems (ITSC), IEEE, 2017, pp. 1--7.

\bibitem{kumar2018near}
V.~R. Kumar, S.~Milz, C.~Witt, M.~Simon, K.~Amende, J.~Petzold, S.~Yogamani, T.~Pech, Near-field depth estimation using monocular fisheye camera: A semi-supervised learning approach using sparse lidar data, in: CVPR Workshop, Vol.~7, 2018, p.~2.

\bibitem{kumar2021svdistnet}
V.~R. Kumar, M.~Klingner, S.~Yogamani, M.~Bach, S.~Milz, T.~Fingscheidt, P.~M{\"a}der, Svdistnet: Self-supervised near-field distance estimation on surround view fisheye cameras, IEEE Transactions on Intelligent Transportation Systems 23~(8) (2021) 10252--10261.

\bibitem{sekkat2022synwoodscape}
A.~R. Sekkat, Y.~Dupuis, V.~R. Kumar, H.~Rashed, S.~Yogamani, P.~Vasseur, P.~Honeine, Synwoodscape: Synthetic surround-view fisheye camera dataset for autonomous driving, IEEE Robotics and Automation Letters 7~(3) (2022) 8502--8509.

\bibitem{kumar2020unrectdepthnet}
V.~R. Kumar, S.~Yogamani, M.~Bach, C.~Witt, S.~Milz, P.~M{\"a}der, Unrectdepthnet: Self-supervised monocular depth estimation using a generic framework for handling common camera distortion models, in: 2020 IEEE/RSJ International Conference on Intelligent Robots and Systems (IROS), IEEE, 2020, pp. 8177--8183.

\bibitem{dhananjaya2021weather}
M.~M. Dhananjaya, V.~R. Kumar, S.~Yogamani, Weather and light level classification for autonomous driving: Dataset, baseline and active learning, in: 2021 IEEE International Intelligent Transportation Systems Conference (ITSC), IEEE, 2021, pp. 2816--2821.

\bibitem{uricar2019desoiling}
M.~Uric{\'a}r, J.~Ulicny, G.~Sistu, H.~Rashed, P.~Krizek, D.~Hurych, A.~Vobecky, S.~Yogamani, Desoiling dataset: Restoring soiled areas on automotive fisheye cameras, in: Proceedings of the IEEE/CVF International Conference on Computer Vision Workshops, 2019, pp. 0--0.

\bibitem{uricar2019challenges}
M.~Uric{\'a}r, D.~Hurych, P.~Krizek, et~al., {Challenges in designing datasets and validation for autonomous driving}, in: Proceedings of the International Conference on Computer Vision Theory and Applications, 2019.

\bibitem{ramachandran2021woodscape}
S.~Ramachandran, G.~Sistu, J.~McDonald, S.~Yogamani, Woodscape fisheye semantic segmentation for autonomous driving--cvpr 2021 omnicv workshop challenge, arXiv preprint arXiv:2107.08246.

\bibitem{siam2018modnet}
M.~Siam, H.~Mahgoub, M.~Zahran, S.~Yogamani, M.~Jagersand, A.~El-Sallab, Modnet: Moving object detection network with motion and appearance for autonomous driving, arXiv preprint arXiv:1709.04821.

\bibitem{rashed2019motion}
H.~Rashed, A.~El~Sallab, S.~Yogamani, M.~ElHelw, Motion and depth augmented semantic segmentation for autonomous navigation, in: Proceedings of the IEEE/CVF Conference on Computer Vision and Pattern Recognition Workshops, 2019, pp. 0--0.

\bibitem{mohamed2021monocular}
E.~Mohamed, M.~Ewaisha, M.~Siam, H.~Rashed, S.~Yogamani, W.~Hamdy, M.~El-Dakdouky, A.~El-Sallab, Monocular instance motion segmentation for autonomous driving: Kitti instancemotseg dataset and multi-task baseline, in: 2021 IEEE Intelligent Vehicles Symposium (IV), IEEE, 2021, pp. 114--121.

\bibitem{kiran2011improved}
B.~R. Kiran, K.~Ramakrishnan, Y.~S. Kumar, K.~Anoop, An improved connected component labeling by recursive label propagation, in: 2011 National Conference on Communications (NCC), IEEE, 2011, pp. 1--5.

\bibitem{tripathi2020trained}
N.~Tripathi, S.~Yogamani, Trained trajectory based automated parking system using visual slam on surround view cameras, arXiv preprint arXiv:2001.02161.

\bibitem{yahiaoui2019overview}
L.~Yahiaoui, J.~Horgan, B.~Deegan, S.~Yogamani, C.~Hughes, P.~Denny, Overview and empirical analysis of isp parameter tuning for visual perception in autonomous driving, Journal of Imaging 5~(10) (2019) 78.

\bibitem{yahiaoui2011impact}
L.~Yahiaoui, J.~Horgan, S.~Yogamani, C.~Hughes, B.~Deegan, Impact analysis and tuning strategies for camera image signal processing parameters in computer vision, in: Irish Machine Vision and Image Processing conference (IMVIP), Vol.~2, 2011.

\bibitem{leang2020dynamic}
I.~Leang, G.~Sistu, F.~B{\"u}rger, A.~Bursuc, S.~Yogamani, Dynamic task weighting methods for multi-task networks in autonomous driving systems, in: 2020 IEEE 23rd International Conference on Intelligent Transportation Systems (ITSC), IEEE, 2020, pp. 1--8.

\bibitem{sistu2019neurall}
G.~Sistu, I.~Leang, S.~Chennupati, S.~Yogamani, C.~Hughes, S.~Milz, S.~Rawashdeh, Neurall: Towards a unified visual perception model for automated driving, in: 2019 IEEE Intelligent Transportation Systems Conference (ITSC), IEEE, 2019, pp. 796--803.

\bibitem{rashed2019optical}
H.~Rashed, S.~Yogamani, A.~El-Sallab, P.~Krizek, M.~El-Helw, Optical flow augmented semantic segmentation networks for automated driving, arXiv preprint arXiv:1901.07355.

\bibitem{kumar2023surround}
V.~R. Kumar, C.~Eising, C.~Witt, S.~Yogamani, Surround-view fisheye camera perception for automated driving: Overview, survey \& challenges, IEEE Transactions on Intelligent Transportation Systems.

\bibitem{popperli2019capsule}
M.~P{\"o}pperli, R.~Gulagundi, S.~Yogamani, S.~Milz, Capsule neural network based height classification using low-cost automotive ultrasonic sensors, in: 2019 IEEE Intelligent Vehicles Symposium (IV), IEEE, 2019, pp. 661--666.

\bibitem{dasgupta2022spatio}
K.~Dasgupta, A.~Das, S.~Das, U.~Bhattacharya, S.~Yogamani, Spatio-contextual deep network-based multimodal pedestrian detection for autonomous driving, IEEE transactions on intelligent transportation systems 23~(9) (2022) 15940--15950.

\bibitem{eising2021near}
C.~Eising, J.~Horgan, S.~Yogamani, Near-field perception for low-speed vehicle automation using surround-view fisheye cameras, IEEE Transactions on Intelligent Transportation Systems 23~(9) (2021) 13976--13993.

\bibitem{Lan2015ContinuousCP}
X.~Lan, S.~D. Cairano, Continuous curvature path planning for semi-autonomous vehicle maneuvers using rrt, 2015 European Control Conference (ECC) (2015) 2360--2365.

\bibitem{Lau2014RealTimePP}
G.~Lau, H.~H.-T. Liu, Real-time path planning algorithm for autonomous border patrol: Design, simulation, and experimentation, Journal of Intelligent \& Robotic Systems 75 (2014) 517--539.

\bibitem{article3}
X.~Zhang, J.~Chen, B.~Xin, Path planning for unmanned aerial vehicles in surveillance tasks under wind fields, Journal of Central South University 21 (2014) 3079--3091.
\newblock \href {http://dx.doi.org/10.1007/s11771-014-2279-7} {\path{doi:10.1007/s11771-014-2279-7}}.

\bibitem{10.1007/978-3-642-33415-3_58}
N.~Ahmidi, G.~D. Hager, L.~Ishii, G.~L. Gallia, M.~Ishii, Robotic path planning for surgeon skill evaluation in minimally-invasive sinus surgery, in: N.~Ayache, H.~Delingette, P.~Golland, K.~Mori (Eds.), Medical Image Computing and Computer-Assisted Intervention -- MICCAI 2012, Springer Berlin Heidelberg, Berlin, Heidelberg, 2012, pp. 471--478.

\bibitem{naderi2015rt}
K.~Naderi, J.~Rajam{\"a}ki, P.~H{\"a}m{\"a}l{\"a}inen, Rt-rrt* a real-time path planning algorithm based on rrt, in: Proceedings of the 8th ACM SIGGRAPH Conference on Motion in Games, 2015, pp. 113--118.

\bibitem{reif1999nano}
J.~Reif, Z.~Sun, Nano-robotics motion planning and its applications in nanotechnology and biomolecular computing, in: NSF Design and Manufacturing Grantees Conference, 1999, pp. 5--8.

\bibitem{4082128}
P.~E. Hart, N.~J. Nilsson, B.~Raphael, A formal basis for the heuristic determination of minimum cost paths, IEEE Transactions on Systems Science and Cybernetics 4~(2) (1968) 100--107.
\newblock \href {http://dx.doi.org/10.1109/TSSC.1968.300136} {\path{doi:10.1109/TSSC.1968.300136}}.

\bibitem{article1}
K.~Daniel, A.~Nash, S.~Koenig, A.~Felner, Theta*: Any-angle path planning on grids, J. Artif. Intell. Res. (JAIR) 39.
\newblock \href {http://dx.doi.org/10.1613/jair.2994} {\path{doi:10.1613/jair.2994}}.

\bibitem{6722915}
M.~Elbanhawi, M.~Simic, Sampling-based robot motion planning: A review, IEEE Access 2 (2014) 56--77.
\newblock \href {http://dx.doi.org/10.1109/ACCESS.2014.2302442} {\path{doi:10.1109/ACCESS.2014.2302442}}.

\bibitem{Hwang1992GrossMP}
Y.~K. Hwang, N.~Ahuja, Gross motion planning—a survey, ACM Comput. Surv. 24 (1992) 219--291.

\bibitem{Goerzen2010ASO}
C.~L. Goerzen, Z.~Kong, B.~Mettler, A survey of motion planning algorithms from the perspective of autonomous uav guidance, Journal of Intelligent and Robotic Systems 57 (2010) 65--100.

\bibitem{10.1016/j.advengsoft.2014.09.006}
J.-H. Liang, C.-H. Lee, Efficient collision-free path-planning of multiple mobile robots system using efficient artificial bee colony algorithm, Adv. Eng. Softw. 79~(C) (2015) 47–56.
\newblock \href {http://dx.doi.org/10.1016/j.advengsoft.2014.09.006} {\path{doi:10.1016/j.advengsoft.2014.09.006}}.

\bibitem{doi:10.5772/63484}
X.~Zhang, Y.~Zhao, N.~Deng, K.~Guo, Dynamic path planning algorithm for a mobile robot based on visible space and an improved genetic algorithm, International Journal of Advanced Robotic Systems 13~(3) (2016) 91.
\newblock \href {http://dx.doi.org/10.5772/63484} {\path{doi:10.5772/63484}}.

\bibitem{KALA20123817}
R.~Kala, Multi-robot path planning using co-evolutionary genetic programming, Expert Systems with Applications 39~(3) (2012) 3817--3831.
\newblock \href {http://dx.doi.org/https://doi.org/10.1016/j.eswa.2011.09.090} {\path{doi:https://doi.org/10.1016/j.eswa.2011.09.090}}.

\bibitem{ZHU2014153}
W.~Zhu, H.~Duan, Chaotic predator–prey biogeography-based optimization approach for ucav path planning, Aerospace Science and Technology 32~(1) (2014) 153--161.

\bibitem{10.1016/j.neucom.2013.07.055}
X.~Wang, Z.-G. Hou, F.~Lv, M.~Tan, Y.~Wang, \href{https://doi.org/10.1016/j.neucom.2013.07.055}{Mobile robots' modular navigation controller using spiking neural networks}, Neurocomput. 134 (2014) 230–238.
\newblock \href {http://dx.doi.org/10.1016/j.neucom.2013.07.055} {\path{doi:10.1016/j.neucom.2013.07.055}}.
\newline\urlprefix\url{https://doi.org/10.1016/j.neucom.2013.07.055}

\bibitem{Zhu2014ThePP}
D.~Zhu, W.~Li, M.~Yan, S.~X. Yang, The path planning of auv based on d-s information fusion map building and bio-inspired neural network in unknown dynamic environment, International Journal of Advanced Robotic Systems 11.

\bibitem{Nosrati2012InvestigationOT}
M.~S. Nosrati, R.~Karimi, H.~A. Hasanvand, Investigation of the * (star) search algorithms: Characteristics, methods and approaches - ti journals, World Applied Programming.

\bibitem{inproceedings1}
M.~Costa, M.~Silva, A survey on path planning algorithms for mobile robots, 2019, pp. 7--18.
\newblock \href {http://dx.doi.org/10.1109/ICARSC.2019.8733623} {\path{doi:10.1109/ICARSC.2019.8733623}}.

\bibitem{Noreen2016OptimalPP}
I.~Noreen, A.~Khan, Z.~Habib, Optimal path planning using rrt* based approaches: A survey and future directions, International Journal of Advanced Computer Science and Applications 7.

\bibitem{Karaman2011SamplingbasedAF}
S.~Karaman, E.~Frazzoli, Sampling-based algorithms for optimal motion planning, The International Journal of Robotics Research 30 (2011) 846 -- 894.

\bibitem{Castañeda08}
M.~A.~P. Castañeda, J.~Savage, A.~Hernández, F.~A. Cosío, \href{https://doi.org/10.5772/6022}{Local autonomous robot navigation using potential fields}, in: X.-J. Jing (Ed.), Motion Planning, IntechOpen, Rijeka, 2008, Ch.~1.
\newblock \href {http://dx.doi.org/10.5772/6022} {\path{doi:10.5772/6022}}.
\newline\urlprefix\url{https://doi.org/10.5772/6022}

\bibitem{7995717}
F.~Bounini, D.~Gingras, H.~Pollart, D.~Gruyer, Modified artificial potential field method for online path planning applications, in: 2017 IEEE Intelligent Vehicles Symposium (IV), 2017, pp. 180--185.
\newblock \href {http://dx.doi.org/10.1109/IVS.2017.7995717} {\path{doi:10.1109/IVS.2017.7995717}}.

\bibitem{Cho2018ARO}
J.~Cho, D.~S. Pae, M.-T. Lim, T.-K. Kang, A real-time obstacle avoidance method for autonomous vehicles using an obstacle-dependent gaussian potential field, Journal of Advanced Transportation.

\bibitem{mashadi2014global}
B.~Mashadi, M.~Majidi, Global optimal path planning of an autonomous vehicle for overtaking a moving obstacle (2014).

\bibitem{508439}
L.~Kavraki, P.~Svestka, J.-C. Latombe, M.~Overmars, Probabilistic roadmaps for path planning in high-dimensional configuration spaces, IEEE Transactions on Robotics and Automation 12~(4) (1996) 566--580.
\newblock \href {http://dx.doi.org/10.1109/70.508439} {\path{doi:10.1109/70.508439}}.

\bibitem{doi:10.5772/56718}
J.~Nasir, F.~Islam, U.~Malik, Y.~Ayaz, O.~Hasan, M.~Khan, M.~S. Muhammad, Rrt*-smart: A rapid convergence implementation of rrt*, International Journal of Advanced Robotic Systems 10~(7) (2013) 299.
\newblock \href {http://dx.doi.org/10.5772/56718} {\path{doi:10.5772/56718}}.

\bibitem{8813752}
S.~Sedighi, D.-V. Nguyen, K.-D. Kuhnert, Guided hybrid a-star path planning algorithm for valet parking applications, in: 2019 5th International Conference on Control, Automation and Robotics (ICCAR), 2019, pp. 570--575.
\newblock \href {http://dx.doi.org/10.1109/ICCAR.2019.8813752} {\path{doi:10.1109/ICCAR.2019.8813752}}.

\bibitem{Dosovitskiy17}
A.~Dosovitskiy, G.~Ros, F.~Codevilla, A.~Lopez, V.~Koltun, {CARLA}: {An} open urban driving simulator, in: Proceedings of the 1st Annual Conference on Robot Learning, 2017, pp. 1--16.

\end{thebibliography}











\clearpage







\end{document}